\documentclass{article}
\usepackage{times}  
\usepackage{helvet} 
\usepackage{courier}  
\usepackage[hyphens]{url}  
\usepackage{graphicx} 
\usepackage{graphicx}  
\usepackage{graphicx}
\usepackage{amsmath}
\usepackage{amsfonts}
\usepackage{dsfont}
\usepackage{amssymb}
\usepackage{lineno}
\usepackage{xcolor,colortbl}
\usepackage{caption}
\usepackage{subfig}
\usepackage{rotating}
\usepackage{hyperref}
\usepackage{algorithm}
\usepackage[noend]{algpseudocode}

\usepackage{subfig}

\DeclareMathOperator*{\argmin}{\arg\!\min}

\begin{document}
\title{Sequential Training of Neural Networks with Gradient Boosting\thanks{This paper is under consideration at Pattern Recognition Letters.}}
\author{Seyedsaman Emami, Gonzalo Mart\'{\i}nez-Mu\~noz}
\date{Escuela Polit\'ectica Superior, Universidad Aut\'onoma de Madrid, Spain}

\maketitle

\begin{abstract}
This paper presents a novel technique based on gradient boosting to train the
final layers of a neural network (NN). Gradient boosting is an additive expansion
algorithm in which a series of models are trained sequentially to approximate a
given function. A neural network can also be seen as an additive expansion where the
scalar product of the responses of the last hidden layer and its weights provide
the final output of the network. Instead of training the network as a whole, the
proposed algorithm trains the network sequentially in $T$ steps. First, the bias
term of the network is initialized with a constant approximation that minimizes
the average loss of the data. Then, at each step, a portion of the network,
composed of $J$ neurons, is trained to approximate the pseudo-residuals on the
training data computed from the previous iterations. Finally, the $T$ partial
models and bias are integrated as a single NN with $T \times J$ neurons in the
hidden layer. Extensive experiments in classification and regression tasks, as
well as in combination with deep neural networks, are carried out showing a
competitive generalization performance with respect to neural networks trained
with different standard solvers, such as Adam, L-BFGS, SGD and deep models.
Furthermore, we show that the proposed method design permits to switch off
a number of hidden units during test (the units that were last trained) without
a significant reduction of its generalization ability. This permits the
adaptation of the model to different classification speed requirements on the
fly.

\end{abstract}

\section{Introduction}

Machine learning is becoming a fundamental piece for the success of more and
more applications every day. Some examples of novel applications include
bioactive molecule prediction \cite{babajide_2016_molecule}, renewable energy
prediction \cite{torres_2017_windsolar}, classification of galactic sources
\cite{mirabal_2016_galactic}, or agriculture area for mapping soil contamination
\cite{jia2021mapping}. It is of capital importance to find algorithms that can
efficiently handle complex data. 
Ensemble methods are very effective at improving the generalization accuracy of
multiple simple models  \cite{classifiersarticle,caruana06empirical} or even
complex models such as MLPs \cite{schwenk_2000_boostingnn} or DeepCNNs
\cite{Moghimi2016}. 

In recent years, gradient boosting \cite{gradientboosting,friedman2000additive}, a fairly old
technique has gained much attention, specially due to the novel and
computationally efficient
version of gradient boosting called eXtreme Gradient Boosting or XGBoost
\cite{XGBoost}. Gradient boosting builds a model as an additive expansion of
regressors to gradually minimize a given loss function. When gradient boosting
is combined with several stochastic techniques, as bootstrapping or feature
sampling from random forest \cite{randomforests}, its performance generally
improves \cite{stochastic_gb}. In fact, this combination of randomization
techniques and optimization has placed XGBoost among the top contenders in
Kaggle competitions \cite{XGBoost} and provides excellent performance in a
variety of applications as in the ones mentioned above. Based on the success of
XGBoost, other techniques have been proposed like CatBoost
\cite{prokhorenkova2018catboost} and LightGBM \cite{ke2017lightgbm}, which
propose improvements in training speed and generalization performance. 
More details
about these methods can be seen in the comparative analysis of Bentéjac
et al. \cite{bentejac_2020_comparative}.
Other type of widespread boosting algorithm is AdaBoost
\cite{freund1997decision}, initially developed for binary classification and
then for multi-class classification (AdaBoost-SAMME)
\cite{hastie2009multi} and regression \cite{drucker1997improving}. 

On the other hand, convolutional deep architectures have shown outstanding
performances especially with structured data such as images, speech, etc.
\cite{lecun2015deep,schmidhuber2015deep}. However, in the context of tabular data,
ensembles of classifiers or simple MLPs are generally more effective than
convolutional deep neural networks \cite{comparison2017zhang}. The objective of this study is to combine the stage-wise optimization of
gradient boosting into the training procedure of the last layers of a
neural network. The
result of the proposed algorithm is an alternative for training a single neural
network (not an ensemble of networks).

Several related studies propose hybrid algorithms that, for instance, transform
a decision forest into a single neural network \cite{welbl2014casting,Biau2018}
or that use a deep architecture to train a tree forest
\cite{Kontschieder_2015_ICCV}. In \cite{welbl2014casting},
it is shown that a pre-trained tree forest can be cast into a two-layer neural
network with the same predictive outputs. First, each tree is converted into a
neural network. To do so, each split in the tree is transformed into an
individual neuron that is connected to a single input attribute (split
attribute) and whose activation threshold is set to the split threshold. In this
way, and by a proper combination of the outputs of these neurons (splits) the
network mimics the behavior of the decision tree. Finally, all neurons are
combined through a second layer, which recovers the forest decision. The weights
of this network can be later retrained to obtain further improvements
\cite{Biau2018}. In \cite{Kontschieder_2015_ICCV}, a
decision forest is trained jointly by means of a deep neural network that learns
all splits of all trees of the forest. To guide the network to learn the splits
of the trees, a procedure that trains the trees using back-propagation, is
proposed. The final output of the algorithm is a decision forest whose
performance is remarkable in image classification tasks.

In other related line of work \cite{nitanda2018functional,huang2018learning},
boosting is applied to the construction of Deep Residual Learning models
\cite{he2016deep}. In \cite{nitanda2018functional}, a novel ResNet weight
estimation model is proposed by generalizing the boosting functional gradient
minimization \cite{mason1999boosting} to the feature extraction space of the
network. The work presented in \cite{huang2018learning} also builds
layer-by-layer a ResNet boosting over features, however, it is based on a
different boosting framework \cite{freund1997decision}.  The model (called
BoostResNet) works by learning a linear classifier on the output of each
residual network block to build an ensemble of shallow blocks. One important
advantage of BoostResNet over standard ResNet is its lower computational
complexity although the reported performance is not consistently better with
respect to ResNet. In contrast, the current proposal method, based on
\cite{gradientboosting}, builds a simple shallow network in-width rather than
complex model in-depth and shows very good performance in tabular datasets with
respect to standard back-propagation training methods. Furthermore, our proposal
can adapt on the fly to the use of a reduced number of hidden neurons.

Another model that resembles the idea proposed in this paper --yet with a 
different optimization process, final model and objective-- was presented in
\cite{bengio2005convex}. 
They propose a convex optimization algorithm for training a neural network that
theoreticallly could reach the global optimum although its exact implementation
is only feasible for a very low number of input features. In
order to reach the global optimum they control the number of hidden neurons of
the model by adding one neuron at a time to the network and by
including a $L^1$ regularization on the top layer. The proposed idea is a
stepwise algorithm as all weights of the network are optimized at each
iteration. This is done in three optimization steps. First, a new neuron (i.e.
linear model) is added and trained on a weighted loss function similarly to
Adaboost. This weighted loss can only be solved exactly for a very low number of
input features. Then, the output layer, and potentially all input weights, are
optimized using the proposed convex formulation. Finally, the output weights are
regularized to reduce the complexity of the network. This final step sets to
zero some of the output weights to effectively remove the corresponding neurons.
The algorithm is tested on one simple 2D problem in order to assess the validity
of the global optimum approach.

In this paper, we propose a combination of ensembles and neural networks that is
somehow complementary to the work of
\cite{Kontschieder_2015_ICCV}, which is a single neural
network that is trained using an ensemble training algorithm.
Specifically, we propose to train a neural network iteratively as an additive
expansion of simpler models. The algorithm is equivalent to gradient boosting:
first, a constant approximation is computed (assigned to the bias term of the
neural network), then at each step, a regression neural network with a single
(or very few) neuron(s) in the hidden layer is trained to fit the residuals of
the previous models. All these models are then combined to form a single neural
network with one hidden layer. This training procedure provides an alternative
to standard training solvers (such as Adam, L-BFGS or SGD) for training a neural
network.
Other works related to the optimization and convergence of Adam have been
recently proposed
\cite{reddi2019convergence,defazio2021adaptivity,chen2018closing}. They showed
that its convergence can be improved in the context of high dimensional complex
image classification tasks. However, their focus is mainly in deep models for
image classification tasks. 

In addition, the proposed method has an additive neural architecture in
which the latest computed neurons contribute less to the final decision. This
can be useful in computationally intensive applications as the number of active
models (or neurons) can be gauged on the fly to the available computational
resources without a significant loss in generalization accuracy. The proposed
model is tested on multiple classification and regression problems, as well as
in conjunction with deep models posed as transfer learning problems. 
These experiments show that the proposed method for training the last layers of
a neural network is a good alternative to other standard methods.

The paper is organized as follows: Section 2 describes gradient boosting and how
to apply it to train a single neural network; In section 3 the results of
several experimental analysis are shown; Finally, the conclusions are summarized
in the last section.

\section{Methodology}
In this section, we show the gradient boosting mathematical framework
\cite{gradientboosting} and the applied modifications in order to use it for
training a neural network sequentially. The proposed algorithm is valid for
multi-class and binary classification, and for regression. Finally, an
illustrative example is given.

\subsection{Gradient boosting}
\label{sec:hyper}
Given a training dataset $D=\{\mathbf{x}_i,y_i\}_{1}^N$, the goal of machine
learning algorithms is to find an approximation, $\hat{F}(\mathbf{x})$, of the
objective function $F^*(\mathbf{x})$, which maps instances $\mathbf{x}$ to their
output values $y$. In general, the learning process can be posed as an
optimization problem in which the expected value of a given loss function,
$\mathds{E}[L(y, F(\mathbf{x}))]$, is minimized. A data-based estimate can be
used to approximate this expected loss: $\sum_{i=1}^N L(y_i,F(\mathbf{x}_i))$.

In the specific case of gradient boosting, the model is built using an additive
expansion 
\begin{equation}
F_t(\mathbf{x}) = F_{t-1}(\mathbf{x}) + \rho_t h_t(\mathbf{x}),
\label{eq:modelexp}
\end{equation}
where $\rho_t$ is the weight of the $t^{th}$ function, $h_t(\mathbf{x})$. The
approximation is constructed {\it stage-wise} in the sense that at each step, a
new model $h_t$ is built without modifying any of the previously created models
included in $F_{t-1}(\mathbf{x})$. First, the additive expansion is initialized with
a constant approximation 
\begin{equation}
F_0(\mathbf{x}) = \argmin_{\alpha} \sum_{i=1}^N L(y_i, \alpha) \;
\label{eq:f0}
\end{equation}
and the following models are built in order to minimize
\begin{equation}
(\rho_t, h_t(\mathbf{x})) = \argmin_{\rho,h_t} \sum_{i=1}^N L(y_i, F_{t-1}(\mathbf{x}_i) + \rho h_t(\mathbf{x}_i)) \;\; .
\end{equation}
However, instead of jointly solve the optimization for $\rho $ and $h_t$, the
problem is split into two steps. First, each model $h_t$ is trained to learn the
data-based gradient vector of the loss-function. For that, each model, $h_t$, is
trained on a new dataset $D=\{\mathbf{x}_i,r_{ti}\}_{i=1}^N$, where the
pseudo-residuals, $r_{ti}$, are the negative gradient of the loss function at
$F_{t-1 (\mathbf{x_i})}$ 
\begin{equation}
r_{ti} =  \left.  - \frac{\partial L(y_i,F(\mathbf{x}_i))}{\partial F(\mathbf{x}_i)} \right|_{F(\mathbf{x})=F_{t-1}(\mathbf{x})}
\label{eq:pseudor}
\end{equation}
The function, $h_t$, is expected to output values close to the pseudo residuals
at the given data points, which are parallel to the gradient of $L$ at
$F_{t-1}(\mathbf{x})$. Note, however, that the training process of $h$ is
generally guided by square-error loss, which may be different from the given
objective loss function. Notwithstanding, the value of $\rho_t$ is subsequently
computed by solving a line search optimization problem on the given loss
function
\begin{equation}
\rho_t = \argmin_{\rho} \sum_{i=1}^N L(y_i,F_{t-1}(\mathbf{x}_i) + \rho h_t(\mathbf{x}_i)) \;\; .
\label{eq:line_s}
\end{equation}

The original formulation of gradient boosting (as given in 
\cite{gradientboosting}) is, in some final derivations,
only valid for decision trees. Here, we present an extension of the formulation
of gradient boosting to be able to use any possible regressor as base model and
we describe how to integrate this process to train a single neural network,
which is the focus of the paper.

\subsection{Binary classification}
For binary classification, in which $y \in \{-1,1\}$, we will consider the logistic loss
\begin{equation}
L(y, F(\cdot)) = \ln{(1+\exp{(-2 y F(\cdot))})} \; ,
\label{eq:logloss}
\end{equation}
which is optimized by the logit function $F(\mathbf{x})=\frac{1}{2} \ln{\frac{p(y=1|\mathbf{x})}{p(y=-1|\mathbf{x})}}$. For this loss function the constant approximation of Eq.~\ref{eq:f0} is given by
\begin{align*}
F_0 &= \argmin_{\alpha} \sum_{i=1}^N \ln(1+\exp(-2 y_i \alpha)) = \\
    &= \frac{1}{2}\ln\frac{p(y=1)}{p(y=-1)}  =
\frac{1}{2}\ln\frac{1-\overline{y}}{1+\overline{y}} \;\;,
\end{align*}
where $\overline{y}$ is the mean value of the class labels $y_i$. The pseudo-residuals given by Eq.\ref{eq:pseudor} on which the model $h_t$ is trained for the logistic loss can be calculated as
\begin{equation}
r_{ti} = 2 y_i / (1 + \exp{(2 y_i F_{t-1}(\mathbf{x}_i))}) \;\;.
\label{eq:res}
\end{equation}
Once $h_t$ is built, the value of $\rho_t$ is computed using Eq.~\ref{eq:line_s} by minimizing
\begin{equation*}
f(\rho) = \sum_{i=1}^N ln(1+\exp(-2y_i(F_{t-1}(\mathbf{x}_i) + \rho h_t(\mathbf{x}_i)))) \;\; .
\end{equation*}
There is no close form solution for this equation. However, the value of $\rho$ can be approximated by a single Newton-Raphson step
\begin{equation}
\rho_t \approx - \frac{f'(\rho=0)}{f''(\rho=0)} = \frac{\sum_{i=1}^{N} r_{ti} h_t({\mathbf{x}}_i)}{\sum_{i=1}^{N} r_{ti} (2 y_i - r_{ti}) h^2_t({\mathbf{x}}_i)} \;\;.
\label{eq:rho}
\end{equation}
This equation is valid for any base regressor and not
only for decision trees as in the general gradient boosting framework
\cite{gradientboosting}. The formulation in \cite{gradientboosting} is adapted
to the fact that decision trees can be seen as piecewise-constant additive
models, which allows for the use of a different $\rho$ for each tree leaf.

Finally, the output for binary classification of gradient boosting composed of $T$ models for instance $\mathbf{x}$ is given by the probability of $y=1|\mathbf{x}$
\begin{equation}
p(y=1|\mathbf{x}) = 1 / (1 + \exp(-2 F_T(\mathbf{x}))) \;\;.
\label{eq:out}
\end{equation}

\subsection{Multi-class classification}

For multi-class classification with $K>2$ classes, the labels are defined with
1-of-K vectors, $\mathbf{y}$, such that $y_{k} = 1$ if the instance belongs to
class $k$ and $y_{k} = 0$, otherwise. In this context the output for
$\mathbf{x}$ of the model is also a $K$-dimensional vector
$\mathbf{F}(\mathbf{x})$. The cross-entropy loss function is used in this
context 
\begin{equation}\label{eq:loglossm}
L(\mathbf{y},\mathbf{F}(\cdot))=-\sum_{k=1}^{K}y_k \ln p_k(\cdot) \;\;,
\end{equation}
where $p_k(\cdot)$ is the probability of a given instance of being of class $k$
\begin{equation}\label{eq:pik}
p_{k}(\cdot)=\frac{\exp(F_k(\cdot))}{\sum_{l=1}^{K}\exp(F_l(\cdot))}
\end{equation}
The additive model is initialized with constant value $0$ as $F_{k,0}=0 \; \forall k$, that correspond to a probability equal to $1/K$ for all classes and instances.

The pseudo-residuals, given by Eq.~\ref{eq:pseudor}, on which the model $\mathbf{h}_t$ is trained for the multi-class loss are the derivative of Eq.~\ref{eq:loglossm} with respect to $F_k$ evaluated at $F_{k,t-1}$
\begin{align}\label{eq:pseudorm}
r_{tik} &= \left. \sum_{j=1}^K \frac{y_{ij}}{p_j(\mathbf{x}_i)} \frac{\partial
p_j(\mathbf{x}_i)}{\partial F_k(\mathbf{x}_i)}
\right|_{F_k(\mathbf{x})=F_{k,t-1}(\mathbf{x})}  \\
&=  \sum_{j=1}^K y_{ij} (\delta_{kj}
- p_{k,t-1}(\mathbf{x}_i)) =  y_{ik}-p_{k,t-1}(\mathbf{x}_i) \;\;,
\end{align}
where $\delta_{kj}$ is Kronecker delta and the fact that $\sum_{j=1}^K y_{ij} = 1 \; \forall i$ is used in the final step. In this study, a single model, $\mathbf{h}_t$ is going to be trained per iteration to fit the residuals for all $K$ classes. In contrast to the $K$ decision trees per iteration that are built in gradient boosting. Then a line search for each of the $K$ outputs of the model is computed by minimizing
\begin{equation}\label{eq:loglossm1}
f(\rho_k) =- \sum_{i=1}^{N}\sum_{j=1}^{K}y_{ij}\ln \left[
\frac{\exp(F_{j,t-1}(\mathbf{x}_i)+\rho_j
h_{j,t}(\mathbf{x}_i))}{\sum_{l=1}^{K}\exp(F_{l,t-1}(\mathbf{x}_i)+\rho_l
h_{l,t}(\mathbf{x}_i))}\right]
\end{equation}
with a Newton-Raphson step
\begin{equation}\label{eq:newton}
\rho_{k,t} = - \frac{f'(\rho _k=0)}{f'' (\rho _k=0)}= -\left[
\frac{\sum_{i=1}^{N} h_{k,t}(\mathbf{x}_i)(y_{ik}-p_{k,t-1}(\mathbf{x}_i))}
{\sum_{i=1}^{N}
h_{k,t}^2(\mathbf{x}_i)p_{k,t-1}(\mathbf{x}_i)(p_{k,t-1}(\mathbf{x}_i)-1)}\right]
\;\; .
\end{equation}
In the same way as for Eq.~\ref{eq:rho}, this equation is valid for all types of base learners and not specific for decision trees as in the original formulation \cite{gradientboosting}, which opens the possibility to apply gradient boosting to other base learners.

The final output of gradient boosting composed of T models for multi class tasks is the probability $y_k=1|\mathbf{x}$
\begin{equation}\label{eq:outm}
p(y_k=1|\mathbf{x}) = \frac{\exp (F_{k,T} ( \mathbf{x}_i))} {\sum_{l=1}^{K}\exp (F_{l,T} (\mathbf{x}_i))}
\end{equation}

\begin{figure*}
\centering
\begin{tabular}{@{}c@{\hspace{2cm}}c@{}}
\includegraphics[width=0.45\textwidth]{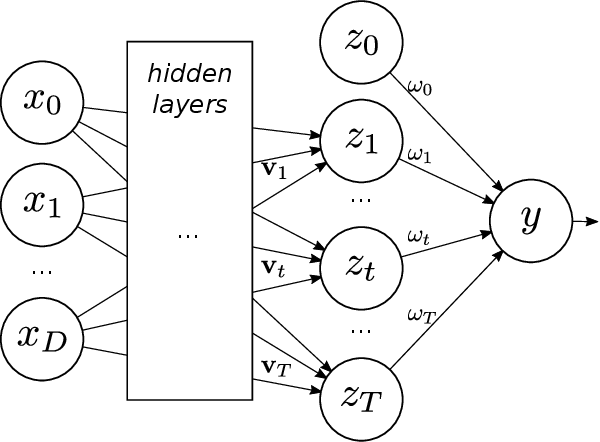}&
\includegraphics[width=0.45\textwidth]{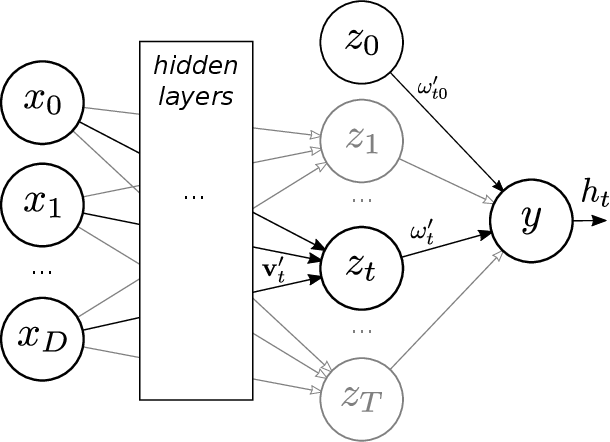}  \\
\includegraphics[width=0.45\textwidth]{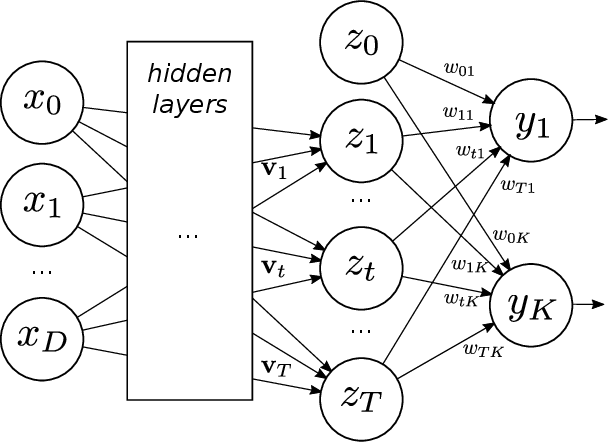}&
\includegraphics[width=0.45\textwidth]{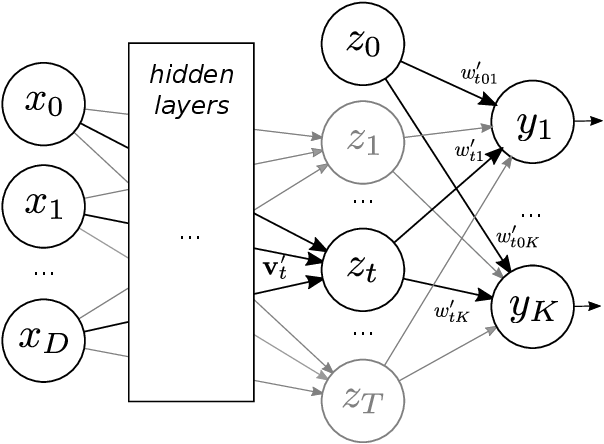}  \\
\end{tabular}
\caption{Illustration of binary and multi-class neural networks and their
parameters (left diagrams) and neural networks with one unit highlighted in
black that represents the $t^{th}$ model trained in gradient boosted neural
network (right diagrams).} \label{fig:gne}
\end{figure*}

\subsection{Neural network as an additive expansion}
A multi-layered neural network can be seen as an additive expansion of its last
hidden layer. The output of the last hidden layer for a fully connected neural
network for binary tasks is (using the parametrization shown in
Fig~\ref{fig:gne} top left) 
\begin{equation}
p(y=1|\mathbf{x})=\sigma\left(\sum_{t=0}^T \omega_t z_t\right)
\label{eq:nn}
\end{equation}
and for multi-class (bottom left in Fig~\ref{fig:gne})
\begin{equation}
p(y=k|\mathbf{x})=\sigma\left(\sum_{t=0}^T \omega_{tk} z_t\right)
\label{eq:nnmc}
\end{equation}
with $z_t$ being the outputs of the last hidden layer, $\omega_t$ and
$\omega_{tk}$ the weights of the last layer and $\sigma$ the activation function
(for classification). For regression, no activation function is used. It is not
straightforward to adapt this process to train deeper layers as we take
advantage of the additive nature of the final output of the network. In the
standard NN training procedure all parameters of the model (i.e. $\mathbf{v}_t$
and $\omega_k$ or $\omega_{tk}$ as shown in Figure~\ref{fig:gne} left diagrams)
are fitted jointly with back-propagation. In the proposed method, instead of
learning the weights jointly, the parameters are trained sequentially using
gradient boosting. To do so, after computing $F_0$, a fully connected regression
neural network with a single (or few) neuron(s) in the last hidden layer is
trained using standard back-propagation. This neural network is trained on the
residuals given by the previous
iteration as given by Eq.~\ref{eq:res} for binary classification and
Eq.~\ref{eq:pseudorm} for multi-class classification tasks. 
Note that, this network trained on the residuals could have a single or several
units in the hidden layer. In the remainder of this section we will assume that
at each step of the boosting procedure a network with one single unit in the
hidden layer is used. Generalizing this to larger networks is trivial.
Figure~\ref{fig:gne}
(right diagrams) shows, highlighted in black, the regression neural network with
a single neuron in the hidden layer that is trained in iteration $t$ and that
corresponds to model $h_t$. 
After model $t$ has been trained, the value of $\rho_t$ is computed using
Eq.~\ref{eq:rho} for binary problems and $\rho_{k,t}$ (Eq.~\ref{eq:newton}) for
multi-class classification. Once all $T$ models have been trained, a neural
network, as shown in Figure~\ref{fig:gne} (left diagrams), with $T$ units in the
last hidden layer is obtained by assigning all the weights necessary to compute
the $z_t$ variables (i.e. ${\mathbf{v}_t}'$ in Figure~\ref{fig:gne} right) to
the corresponding weights in the final NN (i.e. ${\mathbf{v}_t}$ in
Figure~\ref{fig:gne} left) for binary and multi-class tasks 
\begin{align}
\mathbf{v}_t = \mathbf{v}_t' \;\;\;\;  t=1,\dots,T
\label{eq:asig1}
\end{align}
and the weights $\omega_t$ of the output layer for binary classification are
assigned to
\begin{align}
\omega_0 &= F_0 + \sum_{t=1}^T 2 \rho_t {\omega'}_{t0}\\
\omega_t &= 2 \rho_t \omega'_t \;\;\;\;  t=1,\dots,T
\label{eq:asig2}
\end{align}
where the ${\omega'}_{t}$ and ${\omega'}_{t0}$ are the weights from the hidden
neuron to the output and the bias term respectively for the $h_t$ model (as
shown in Figure~\ref{fig:gne} right-top diagram). 
For multi-class with $K$ classes final the assignment is
\begin{align}
\omega_{0k} &= \sum_{t=1}^T \rho_{k,t} {\omega'}_{t0k}\\
\omega_{tk} &= \rho_{k,t} \omega'_{tk} \;\;\;\;  t=1,\dots,T \;\; k=1,\dots,K
\label{eq:asig3}
\end{align}
where the ${\omega'}_{tk}$ and ${\omega'}_{t0k}$ are the output weights for the
$h_t$ multi-class model (Figure~\ref{fig:gne} right-bottom diagram). 

Finally, to recover the probability of $y=k|\mathbf{x}$ in the output of the NN
as given by Eq.~\ref{eq:out} and Eq.~\ref{eq:outm}, the activation function
should be a sigmoid (i.e. $\sigma(x) = 1/(1+  \exp(-x)$) for binary
classification with the logistic loss (Eq.~\ref{eq:logloss}), and the soft-max
activation function (i.e. $\sigma(\mathbf{x})=exp(x_k)/\sum_k exp(x_k)$) for
multi-class tasks when using the cross-entropy loss (Eq.~\ref{eq:loglossm}).
This training procedure can be easily modified to larger increments of neurons,
so that instead of a single neuron per step (a linear model), a more flexible
model, comprising more units ($J$), can be trained at each iteration. 
The outline of the proposed method is shown in Algorithm~\ref{algo:gbnn}.

The proposed training procedure can be further tuned by applying subsampling
and/or shrinking, as generally used in gradient boosting
\cite{stochastic_gb,gradientboosting,XGBoost}. In shrinking, the additive
expansion process is regularized by multiplying each term $\rho_t h_t$ by a
constant learning rate, $\nu \in (0, 1]$, to prevent overfitting when multiple
models are combined \cite{gradientboosting}. Subsampling consists of
training each model on a random subsample without replacement from the original
training data. Subsampling has shown to improve the performance of gradient
boosting \cite{stochastic_gb}. 

The overall computational time complexity to train the proposed algorithm is the
same as that of a standard NN with the same number of hidden neurons and
training epochs. 
Note that, at each step of the proposed algorithm, one NN with one (or few)
neurons in the hidden layer is trained. Hence, at the end of the process, the
same number of weights updates are performed. Note, however, that the proposed
method is sequential and cannot be easily parallelized. Once the model is
trained, the computational complexity for classifying new instances is
equivalent to that of a standard NN as the generated model {\it is} a standard
neural network. 

\begin{algorithm}[H]
\caption{Training Neural network as an additive expansion}\label{algo:gbnn}
	\hspace*{\algorithmicindent} \textbf{Input}
	\begin{algorithmic}[1]
	
	\State Input data $D=\{\mathbf{x}_i,y_i\}_{1}^N$
	\State Number of neurons $T$
	\State Loss function
	\end{algorithmic}
	\hspace*{\algorithmicindent} \textbf{Training the model}
	\begin{algorithmic}[1]
	  \State Initialize ${\hat{F}}_{0}$
		\For {$t=1$ to $T$}
			\State Compute the pseudo-residual with Eqs.~\ref{eq:res}~\ref{eq:pseudorm}
			($r_{ti}$ for binary and $r_{tik}$ for multi-class classification)
			\State Fit a new regressor network model on the residuals.
			\State Compute the gradient descent step through the Newton-Raphson step
			($\rho_{t}$ for binary, Eq.~\ref{eq:rho} and $\rho_{kt}$ for multi-class classification, Eq.~\ref{eq:newton})
			\State Update the model using Eq.\ref{eq:modelexp} 
		\EndFor
		\State \textbf{end for}
    \State Create the final network by assigning the weights with
    Eqs.~\ref{eq:asig1}~\ref{eq:asig2}~\ref{eq:asig3}

	\end{algorithmic} 
\end{algorithm}

\subsection{Illustrative example}
To illustrate the workings of this algorithm, we show its performance in a toy
classification problem. The toy problem task consists in a 2D version of the
{\it ringnorm} problem \cite{breiman96bias}: where both classes are 2D Gaussian
distribution, one with $(0,0)$ mean and covariance four times the identity
matrix, and the second class with mean value at $(2/\sqrt{2},2/\sqrt{2})$ and
the identity matrix as covariance.

\begin{figure*}[t]
\centering
\begin{tabular}{@{}c@{}}
\includegraphics[width=\textwidth]{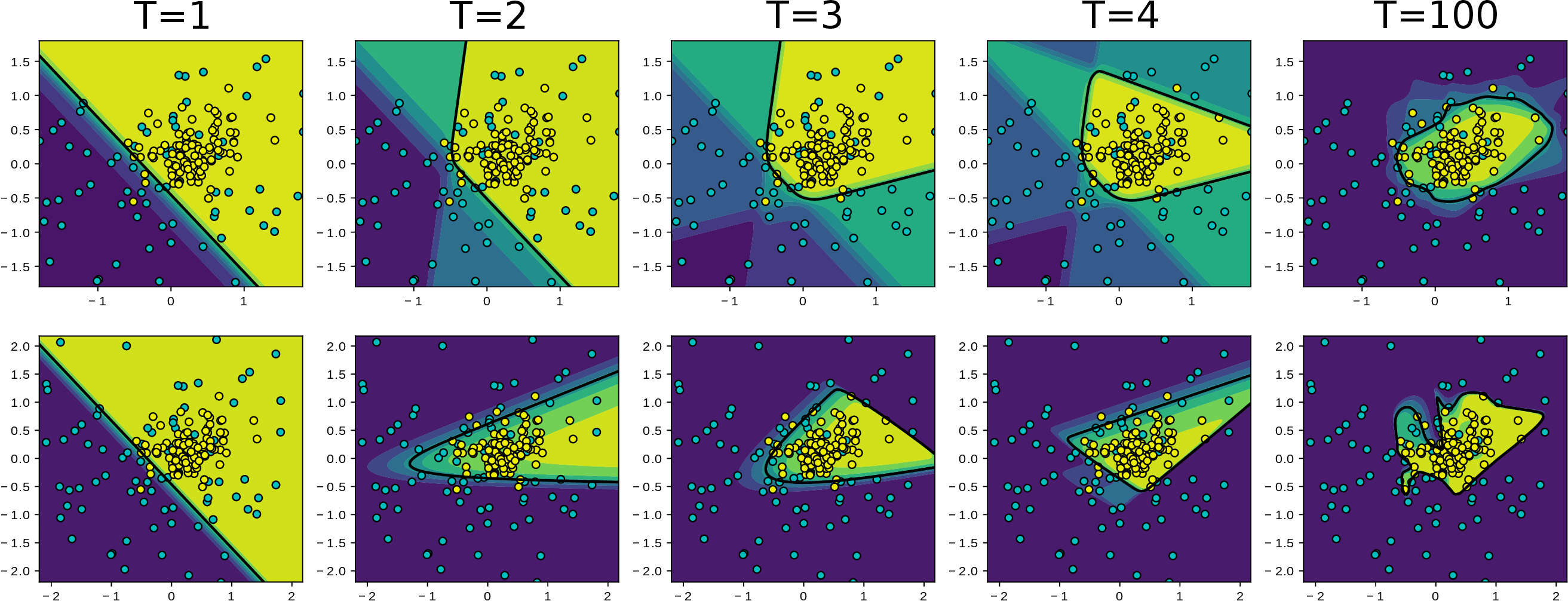}
\end{tabular}
\caption{Classification boundaries for gradient boosted neural network (top row)
and for a standard neural network (bottom row). Each column shows the results
for a combination of a different number of models (hidden units). Top plots are
the sequential results of a single GBNN model, whether bottom plots are
independent neural networks models} \label{fig:simpleexp}
\end{figure*}

The proposed gradient boosted neural network (GBNN) with $T=100$ and one hidden
layer is trained on $200$ randomly generated instances of this problem. In
addition, $100$ independent neural nets with hidden units in the range $[1,100]$
are also trained using the same training set. In Figure~\ref{fig:simpleexp}, the
boundaries for the different stages of the process are shown graphically. In
detail, the first and second rows show the results for GBNN and NN,
respectively. Each column shows the results for $T=1$, $T=2$, $T=3$, $T=4$ and
$T=100$ neurons in the hidden layer respectively. Note that the plots for GBNN
are sequential; That is, the first column shows the first trained model, the
second column, the first two models combined, and so on. For the NN, each column
corresponds to a different NN with a different number of neurons in the hidden
layer. For each column, the architecture of the networks and the number of
weights (but not their values) are the same. The color of the plots represent
the probability $p(y=1\mid \mathbf{x})$ given by the models (Eq.~\ref{eq:out}) using
the {\it viridis} colormap. In addition, all plots show the training points.

As we can see both GBNN and NN start, as expected, with very similar models
(column $T=1$). As the number of models (neurons for NN) increases, GBNN builds
up the boundary from previous models. On the other hand, the standard neural
network, as it creates a new model for each size, is able to adjust faster to
the data. However, as the number of neurons increases, NN tends to overfit in
this problem (as shown in the bottom right-most plot). On the contrary, GBNN
tends to focus on the {\it unsolved} parts of the problem: the decision boundary
becomes defined only asymptotically, as the number of models (neurons) becomes
large.

\section{Experimental results}
In this section, an analysis of the efficiency of the proposed neural network
training method based on gradient boosting is tested on twelve binary
classification tasks, eight multi-class problems, seven regression datasets from
the UCI repository \cite{UCI}, and two datasets related to image processing
\cite{deng2012mnist,krizhevsky2010cifar}. These datasets, shown in
Table~\ref{table:datasets}, have different number of instances and attributes
and come from different fields of application. The {\it Energy} dataset has two
different target columns (cooling and heating), so the different algorithms were
executed for both objectives separately. We modified some of the datasets. For
{\it Diabetes} duplicated instances and instances with missing values were
removed. In addition, categorical values were substituted by dummy variables in:
{\it German Credit Data, Hepatitis, Indian Liver Patient, MAGIC} and {\it
Tic-tac-toe}. Finally, CIFAR-10 and MNIST were normalized by dividing the
attributes by 255.

\begin{table*}[t!]
\centering
\resizebox{\textwidth}{!}{%
\begin{tabular}{llcc}
\hline
\textbf{Dataset} 	& \textbf{Area} 	& \textbf{Instances} & \textbf{Attribs.} \\
\hline
{\bf Binary classification}\\
Australian Credit Approval  &  Financial &  690 &  14 \\
German Credit Data   & Financial & 1,000 & 20 \\
Banknote  & Computer & 1,372 &  5 \\
Spambase  & Computer & 4,601 & 57 \\
Tic-tac-toe  & Game & 958 &  9\\
Breast cancer  &  Life & 569 &  32 \\
Diabetes  & Life & 381 & 9 \\
Hepatitis  & Life & 155 &  19 \\
Indian Liver Patient  & Life & 583 &  10 \\
MAGIC Gamma Telescope  & Physical & 19,020 &  11 \\
Ionosphere  & Physical & 351 &  34 \\
Sonar  & Physical &   208 & 60  \\
\hline
{\bf Multi-class classification}\\
Digits  & Computer & 1,797 &  64 \\
Poker Hand & Game& 1,025,010 & 10 \\
Iris  & Life &  150 & 4 \\
Covertype & Life & 581,012 &   54 \\
Vehicle  & Transportation &  946 & 18 \\
Vowel  & Education & 990 &  11 \\
Waveform & Physical & 5,000 & 21   \\
Wine  & Physical & 178 &  13\\
\hline
{\bf Regression}\\
Concrete  &  Physical &  1,030 &  9 \\
Energy  &  Computer &  768 &  8 \\
Power  &  Computer &  9,568 &  4 \\
Boston Housing  &  Business &  506 &  14 \\
Wine quality-red  &  Business &  1,599 &  12 \\
Wine quality-white  &  Business &  4,898 &  12 \\
\hline
{\bf image processing}\\
MNIST  &  handwritten digits &  60,000  &  28x28 \\
CIFAR-10  &  labeled images &  60,000  &  32x32 \\
\hline
\end{tabular}\vspace*{-25pt}
}
\caption{The details of datasets, which used in experimental analysis}
\label{table:datasets}
\end{table*}

Two batches of experiments were carried out: experiments on tabular data and
experiments on image and large datasets. In the first batch, the proposed
method is compared with respect to standard neural networks using different
solvers and with respect to dense deep neural networks. 
In the second batch, a transfer learning approach was followed, training the last
layer of deep models with the proposed method and fully connected NN.
The first experiment is
carried out in all the classification and regression tasks shown in
Table~\ref{table:datasets}, except for Covertype, Poker Hand, MNIST and
CIFAR-10. For these large datasets, the comparison of the proposed method was
carried out with respect to deep dense or convolutional neural networks, depending on
the type of problem. For the first batch of experiments, the {\it scikit-learn}
package \cite{scikit-learn} was used. For the second batch, we adopted
{\it Keras} library \cite{chollet2015keras}. The implementation of the
proposed method (GBNN) is done in python
following the standards of {\it scikit-learn}. The implementation of the
algorithm is available on
\href{https://github.com/GAA-UAM/GBNN/}{https://github.com/GAA-UAM/GBNN/}. 

\subsection{Experiments with tabular data}
For the first set of experiments on tabular regression and classification,
single hidden layer networks were trained using three different standard solvers
(Adam, L-BFGS and SGD), and using the proposed gradient boosting approach. Also,
we considered a three-layer deep dense neural network trained with the Adam
solver. 
Furthermore, AdaBoost \cite{freund1997decision} using small neural networks as
the base models was also included in the comparison (AdaBoost--NN).
Note that, this approach is different to the proposed GBNN. First, for
classification tasks, the base models in Adaboost are classifiers that are
combined by weighted majority voting. Hence, the final model is a collection of
small base classifiers and not a single neural network as in GBNN. Second,
Adaboost is based on modifying the instance weights during training so that
{\it difficult} instances tend to get higher weights. This poses a
difficulty in the training of the neural networks as they do not handle
weighted instances. In order to run Adaboost with neural networks we included a
weighted resampling step prior the training of each individual network. The
sklearn library does not include this functionality in Adaboost. 
The comparison for classification and
regression problems was carried out using 5$\times$10-fold cross-validation in
order to have stable results. For the Waveform dataset we also considered 10
random train-test partitions using 300 instances for training the models and the
rest for test as experiments with this dataset generally consider this
experimental setup \cite{breiman96bias}. All datasets were standardized in train
so that all attributes have zero mean and one variance. The optimum
hyper-parameter setting for each method was estimated using within-train 10-fold
cross-validation. 
For the standard neural network with the standard
solvers, the grid with the number of units in the hidden layer was set to 
[1, 3, 5, 7, 11, 12, 17, 22, 
27, 32, 37, 42, 47, 52, 60,
70, 80, 90, 100, 150, 200]. 
In addition, for the SGD solver, the learning
policies $[Adaptive, Constant]$ were also considered in its grid search. The
rest of the hyper-parameters were set to their default values. Regarding the
three-layer deep network, 100 neurons per layer were used and all the other
hyper-parameters were left to their default values. For GBNN, a sequentially
trained neural network with 200 hidden units is built in steps of $J$ units per
iteration. For this model, the hyper-parameter grid search was carried out using
the following values for binary classification: $[0.1, 0.25, 0.5, 1.0]$ for the
learning rate, $[0.5, 0.75, 1.0]$ for subsample rate and $J \in [1,2,3]$. For
the multi-class and regression problems the hyper-parameter grid was extended a
bit because in some datasets the values at the extremes were always selected.
Hence, the grid for multi-class and regression problems is set to: $[0.025,
0.05, 0.1, 0.5, 1]$ for the learning rate, $[0.25, 0.5, 0.75, 1.0]$ for
subsample rate and $J \in [1,2,3,4]$. Moreover, for the proposed procedure, the
sub-networks are trained using the L-BFGS solver for all datasets. This decision
was made after some preliminary experiments that showed that L-BFGS provided
good results in general for the small networks we are training at each
iteration.
Regarding AdaBoost-NN, a classification/regression one hidden layer perceptron is
set as base estimator. The number of estimators to include in the ensemble and
number of hidden neurons were set respectively to the following pairs:
(200,1), (100,2) and (67,3). For multi-class problems the pair (50, 4) was also
considered. This is done in order to produce a model with a
combined total number of 200 hidden neurons as for previous networks.
Subsequently, the best sets of hyper-parameters obtained in the
grid search for each method were used to train the whole training set, one
standard neural network for each solver, dense deep neural networks, AdaBoost--NN,
and the proposed gradient boosted neural network. Finally, the average
generalization performance of the models was estimated in the left-out test set. 

\begin{figure}[t!]
\centering
\subfloat[Classification]{{\includegraphics[width=6cm]{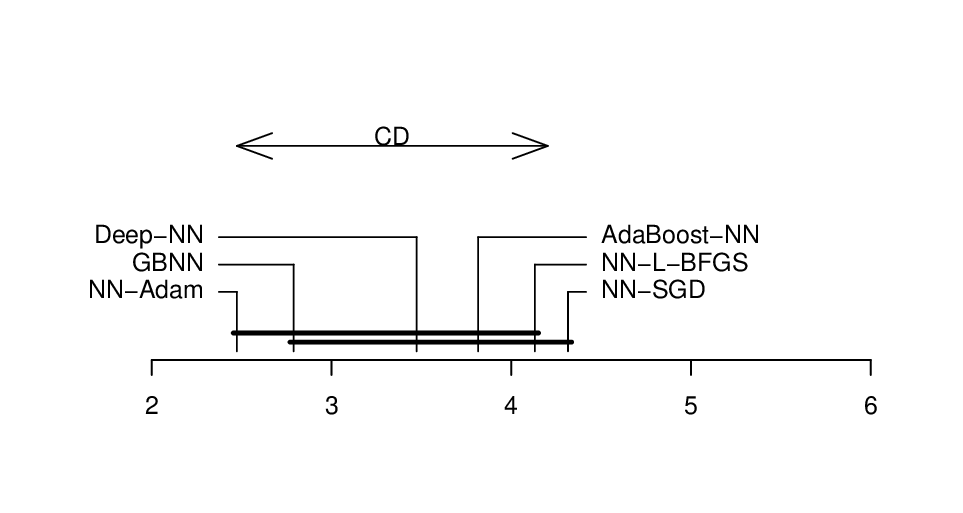}}}
\subfloat[Regression]{{\includegraphics[width=6cm]{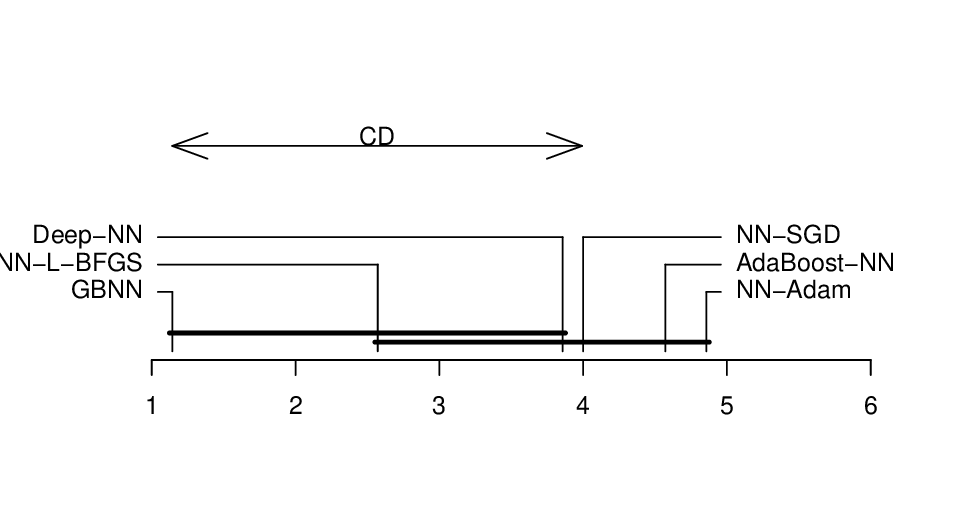}}}
\qquad
\subfloat[All of the datasets]{{\includegraphics[width=6cm]{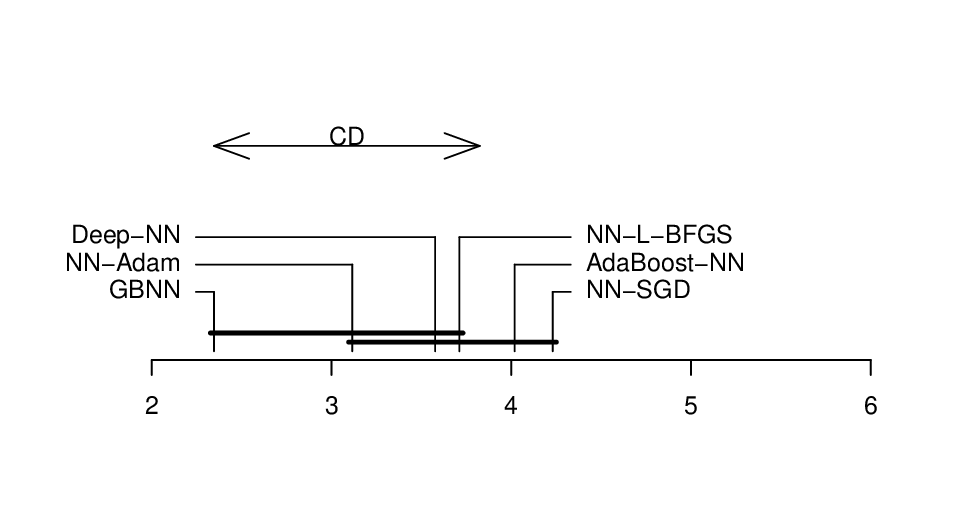}}}
\caption{Average ranks (higher rank is better) for GBNN, Deep-NN, NN--Adam,
NN--L-BFGS, NN--sgd and AdaBoost--NN for 26 datasets. a) The Dems\v{a}r plot for binary and
multiclass datasets. b) The Dems\v{a}r plot for regression datasets. c) The
Dems\v{a}r plot considering all of the datasets.} \label{fig:demsar}
\end{figure}

For all analyzed datasets, the average generalization performance and standard
deviations are shown in Table~\ref{table:results} for the proposed method
(column GBNN), the three-layer deep neural network (column Deep-NN), standard 
neural networks trained with Adam (NN--Adam), L-BFGS (NN--L-BFGS), SGD
(NN--SGD), and AdaBoost--NN. For classification problems the generalization
performance is given as average accuracy and, for regression, as average root
mean square error. The Table~\ref{table:results} is structured in three blocks
depending on the problem type: binary classification tasks, multi-class
classification, and regression. The best result for each method is highlighted
with a light yellow background. An overall comparison of these results is shown
graphically in Figure~\ref{fig:demsar} using the methodology described in
\cite{demsar2006}. These plots show the average rank for the studied methods
across the analyzed datasets where a higher rank (i.e. lower values) indicates
better results. The figure shows the Dems\v{a}r plots for the classifications
tasks only (top left plot), for regression only (top right plot), and for all
analyzed tasks (bottom plot). The statistical differences between methods are
determined using a Nemenyi test. In the plot, the difference in the average rank
of the two methods is statistically significant if the methods are not connected
with a horizontal solid line. The critical distance (CD) in average rank over
which the performance of two methods is considered significant is shown in the
plot for reference (CD = 1.72, 2.84, and 1.47 for 19 classification, 7
regression and all datasets respectively, considering 6 methods and p-value
$<0.05$).

\begin{table*}[t!]
\Large
\centering
\resizebox{0.95\textwidth}{!}{%
\begin{tabular}{lcccccc}
\hline
\textbf{Dataset} & \textbf{GBNN} & \textbf{Deep-NN}	& \textbf{NN--Adam} 	& \textbf{NN--L-BFGS} & \textbf{NN--SGD} & \textbf{AdaBoost--NN} \\
\hline
{\bf Binary classification}\\
Australian Credit Approval &  85.91\%$\pm$4.81 &  82.81\%$\pm$4.26 &  85.88\%$\pm$4.43 & 86.20\%$\pm$4.32 & \cellcolor{yellow!47} 86.35\%$\pm$4.28 &  84.58 \%$\pm$3.32 \\
Banknote & 99.99\%$\pm$0.04 & \cellcolor{yellow!47}100.00\%$\pm$0.00 & 99.99\%$\pm$0.04 &  99.81\%$\pm$0.42  &  97.32\%$\pm$1.15 & \cellcolor{yellow!47} 100.00\%$\pm$0.00 \\
Breast cancer & 96.87\%$\pm$1.93 &  \cellcolor{yellow!47} 97.40\%$\pm$2.06 & 97.36\%$\pm$1.97 &  96.83\%$\pm$1.55 &  96.55\%$\pm$1.69 & 96.45\%$\pm$4.72\\
Diabetes & 76.12\%$\pm$3.93 & 71.30\%$\pm$4.59 & 76.35\%$\pm$4.05 & 76.67\%$\pm$ 4.15 & \cellcolor{yellow!47} 76.96\%$\pm$ 4.21 & 74.81\%$\pm$ 3.25 \\
German Credit Data & \cellcolor{yellow!47}  74.16\%$\pm$3.64 & 69.84\%$\pm$4.53 & 73.74\%$\pm$3.99 & 72.34\%$\pm$ 3.41 & 73.28\%$\pm$ 3.87 & 72.44\%$\pm$ 4.42\\
Hepatitis & 82.81\%$\pm$ 10.60 &  84.10\%$\pm$ 10.83 & 85.10\%$\pm$ 10.86 &  82.11\%$\pm$ 9.21 &  85.09\%$\pm$ 10.65 & \cellcolor{yellow!47} 85.54\%$\pm$ 7.82 \\
Indian Liver Patient & \cellcolor{yellow!47}72.51\%$\pm$5.30 & 70.88\%$\pm$ 5.20 & 70.69\%$\pm$ 5.81 &  69.44\%$\pm$ 5.88 &  71.10\%$\pm$ 5.52 & 69.26\%$\pm$4.84\\
Ionosphere & 90.94\%$\pm$ 4.86 & \cellcolor{yellow!47} 93.28\%$\pm$3.69 & 91.34\%$\pm$3.92 &  90.43\%$\pm$4.58 &  87.90\%$\pm$5.96 & 92.25\%$\pm$3.52\\
MAGIC Gamma Telescope & 87.52\%$\pm$0.67 & 85.62\%$\pm$0.85 & \cellcolor{yellow!47} 87.65\%$\pm$0.57 &  87.58\%$\pm$ 0.62 &  86.28\%$\pm$ 1.48 &  87.19\%$\pm$0.75 \\
Sonar & 78.84\%$\pm$7.30 & \cellcolor{yellow!47} 86.22\%$\pm$7.57 & 85.56\%$\pm$6.48 &   85.29\%$\pm$ 5.07 & 77.67\%$\pm$ 7.67 &  85.91\%$\pm$8.26\\
Spambase &  94.44\%$\pm$1.00 & 94.08\%$\pm$1.22 & \cellcolor{yellow!47} 94.61\%$\pm$1.01 &  93.43\%$\pm$ 1.27 &  93.42\%$\pm$ 1.27 &  74.20\%$\pm$13.16 \\
Tic-tac-toe & \cellcolor{yellow!47}98.70\%$\pm$1.17 & 95.78\%$\pm$1.80 & 90.44\%$\pm$3.19 & 93.53\%$\pm$ 2.56 & 70.94\%$\pm$ 4.30 &  86.75\%$\pm$3.69\\
\hline
{\bf Multi-class classification}\\
Digits & 97.18\%$\pm$1.21 & 97.55\%$\pm$1.06 & \cellcolor{yellow!47}98.04\%$\pm$1.11 &  97.17\%$\pm$1.10 &  96.44\%$\pm$1.09 &  95.66\%$\pm$1.18\\
Iris &\cellcolor{yellow!47} 95.73\%$\pm$ 6.02 & 94.40\%$\pm$6.58 & 95.33\%$\pm$5.58&  94.93\%$\pm$6.57 &  85.07\%$\pm$ 8.71 &  94.72\%$\pm$2.16\\
Vehicle & \cellcolor{yellow!47}84.61\%$\pm$3.92 &  83.41\%$\pm$3.82 & 83.43\%$\pm$3.52 &  82.96\%$\pm$4.21 &  72.99\%$\pm$4.51 &  72.13\%$\pm$4.15\\	
Vowel & 89.88\%$\pm$2.57 & \cellcolor{yellow!47}96.71\%$\pm$1.96 & 94.28\%$\pm$2.36 &  93.13\%$\pm$3.09 &  53.04\%$\pm$4.34 &  51.45\%$\pm$3.37\\
Waveform& \cellcolor{yellow!47}87.00\%$\pm$ 1.16 & 82.60\%$\pm$ 1.62 & 86.65\%$\pm$1.40 &  86.46\%$\pm$ 1.22 &  86.68\%$\pm$ 1.37 & 84.99\%$\pm$1.46\\
Waveform-300& 82.94\%$\pm$ 0.17 & 82.07\%$\pm$ 0.63 & 83.69\%$\pm$0.70 &  79.51\%$\pm$ 0.02 &  83.92\%$\pm$ 0.77 & \cellcolor{yellow!47} 84.55\%$\pm$0.40\\
Wine & \cellcolor{yellow!47} 98.88\%$\pm$ 2.35 & 97.87\%$\pm$ 3.35 & 97.77\%$\pm$ 3.22 &  97.65\%$\pm$3.50 &  95.72\%$\pm$4.91 &  97.32\%$\pm$3.48\\
\hline
{\bf Regression}\\
Boston Housing & \cellcolor{yellow!47} 3.03\%$\pm$0.74 & 3.18\%$\pm$0.79 & 4.12\%$\pm$0.77 &  3.50\%$\pm$0.98 &  3.40\%$\pm$ 0.87&  14.43\%$\pm$1.49\\
Concrete & 4.80\%$\pm$0.59 & 5.20\%$\pm$0.56 & 9,61\%$\pm$0.58 &   \cellcolor{yellow!47}  4.73\%$\pm$0.59 &  6.04\%$\pm$ 0.47&  26.78\%$\pm$1.10\\
Energy-Cooling & \cellcolor{yellow!47} 0.95\%$\pm$0.16 & 2.16\%$\pm$0.31 & 3.45\%$\pm$0.46 &  1.14\%$\pm$0.17 &  3.12\%$\pm$0.41&  15.91\%$\pm$1.69\\
Energy-Heating & \cellcolor{yellow!47} 0.43\%$\pm$0.08 & 1.40\%$\pm$0.27 & 2.96\%$\pm$0.38 &  0.49\%$\pm$0.06 &  2.75\%$\pm$0.33&  12.95\%$\pm$1.29\\
Power   & \cellcolor{yellow!47} 3.84\%$\pm$0.18 & 4.40\%$\pm$0.39 & 4.25\%$\pm$0.16 &  4.11\%$\pm$ 0.17 & 4.14\%$\pm$0.15 & 123.37\%$\pm$9.70\\
Wine quality-red & \cellcolor{yellow!47} 0.60\%$\pm$0.04 & 0.68\%$\pm$0.06 & 0.64\%$\pm$0.05 &  0.64\%$\pm$0.73 &  0.65\%$\pm$0.04 &  0.83\%$\pm$0.06\\ 
Wine quality-white & \cellcolor{yellow!47} 0.67\%$\pm$0.03 & 0.72\%$\pm$0.04 & 0.68\%$\pm$0.03 & 0.70\%$\pm$0.03 &  0.70\%$\pm$0.03&  0.74\%$\pm$0.02\\ 
\hline
\end{tabular}\hspace*{-25pt}
}
\caption{Average generalization performance and standard deviation for gradient
boosted neural network (GBNN), Deep neural network, neural networks trained with
solvers Adam, L-BFGS, SGD and AdaBoost. Using accuracy for the classification
tasks and root mean square error (RMSE) for regression problems as the
performance measurement. The best results for each dataset are highlighted with
a light yellow background.} \label{table:results} 
\end{table*}

From Table~\ref{table:results} and Figure~\ref{fig:demsar}, it can be observed
that, for the studied datasets, the best overall performing method is GBNN. This
method reached the best results
in 13 out of 26 tasks. The deep-NN model performed best in five datasets. The
neural network method with Adam as its solver captured the best results in three
datasets. SGD and L-BFGS solvers obtained the best outcome in two and one
datasets, respectively. And finally, the AdaBoost--NN got the best performance
in three datasets. In classification, the differences in average accuracy among
the different methods are generally favorable to GBNN, NN--Adam and Deep--NN,
although the differences between these methods are small in many datasets. For
instance, in {\it Magic}, the accuracy of GBNN is only 0.13 percent points worse
than NN--Adam. However, small differences are not always the case. One of the
most notable differences between GBNN and second top ranked method is in {\it
Tic-tac-toe} where Deep--NN is 2.92\% worse than the result obtained by GBNN and
NN-Adam is more than 8 points worse. In contrast, the most favorable outcome for
Deep--NN is obtained in {\it Vowel} and {\it Sonar}, where its accuracy is
6.83\% and 7.38\% better than that of GBNN, respectively. The results in
classification for NN–L--BFGS, NN–SGD and AdaBoost--NN are generally worse than
those of the other three methods.

In regression, the results are more clearly in favor of GBNN. The proposed
method obtains the best performance in all tested datasets except for one
dataset, {\it Concrete}, in which GBNN get the second best result. The
performance of NN–Adam in regression is suboptimal getting the worst
performances in general. Solver L-BFGS perform more closely to the performance
of GBNN.

These results can also be observed from Figure~\ref{fig:demsar}. The performance
of NN--L-BFGS, NN--SGD and AdaBoost--NN in the classification tasks is worse
than that of NN--Adam, GBNN and Deep--NN (subplot a). NN-Adam and GBNN take the
highest rank in the statistical test. For regression (subplot b), the
best-performing method is GBNN. The performance of GBNN is significantly better
than that of NN--Adam, NN--SGD and AdaBoost--NN. The overall results (subplot
c) show that GBNN has the best rank followed by NN--Adam and Deep--NN. Overall,
the performance of GBNN is statistically better than the performance of NN--SGD
and AdaBoost-NN.

\begin{figure*}[t!]
\centering
\subfloat[{\it Spambase}]{{\includegraphics[width=4.1cm]{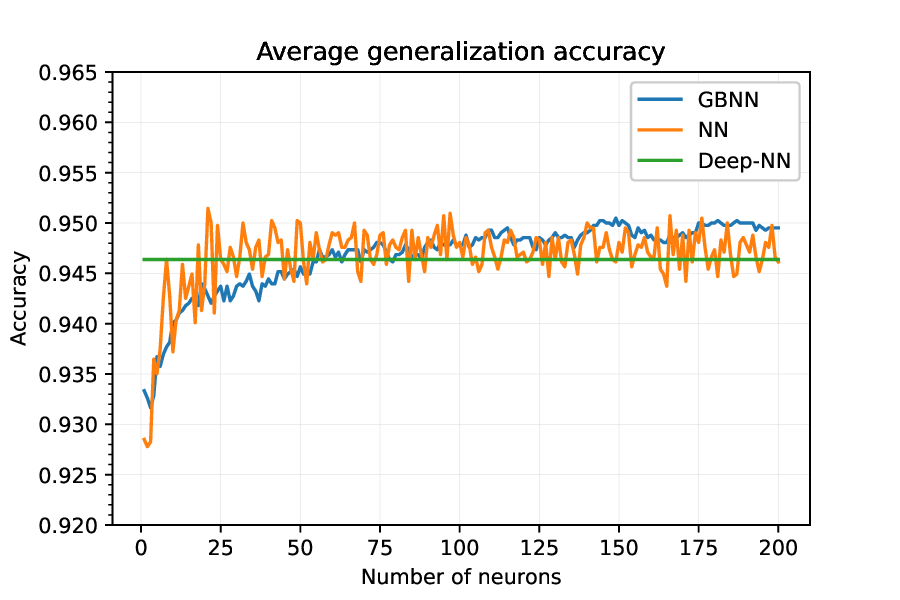}}}
\subfloat[{\it Tic-tac-toe}]{{\includegraphics[width=4.1cm]{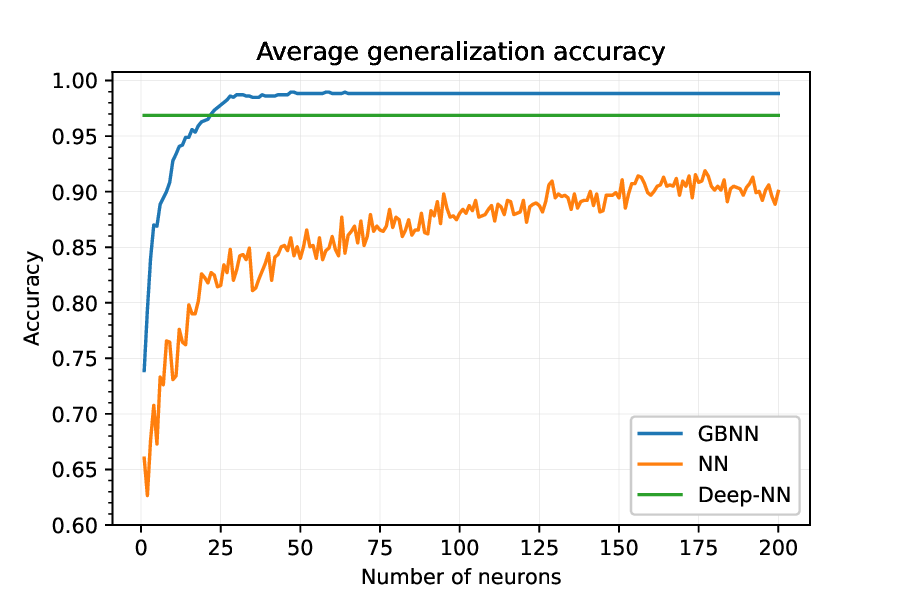}}}
\subfloat[{\it Energy-Heating}]{{\includegraphics[width=4.1cm]{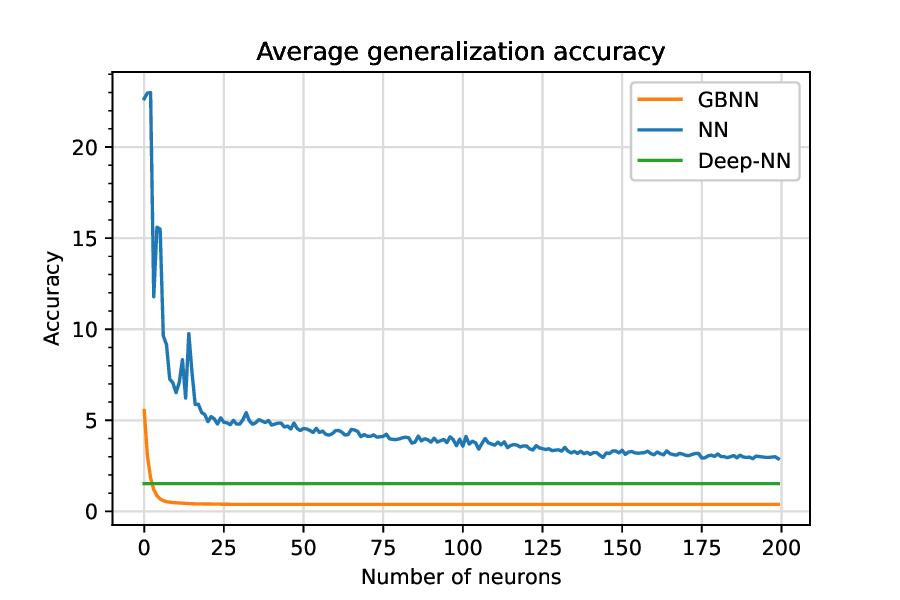}}}
\caption{Average generalization accuracy
for GBNN (blue curve), NN (orange curve), and Deep-NN (green
curve) with respect to the number of hidden units} \label{fig:acc_comparison}
\end{figure*}

To analyze the evolution of the networks generated by the proposed method as
more hidden units are included, another experiment was carried out to build a
GBNN with 200 hidden units, and neural networks trained using the Adam solver
with 1 to 200 neurons in the hidden layer. In addition, the final accuracy of a
three-layer deep neural network with 100 neurons in each layer is also computed
for reference. For this experiment, 10-fold cross-validation was used. 
The average evolution is shown in Figure~\ref{fig:acc_comparison}
for {\it Spambase} (left plot), {\it Tic-tac-toe} (middle plot) and {\it Energy-Heating} (right plot).
Due to computational limitations in this case, the hyper-parameters were not
tuned for each partition. Instead, they were set to the values more often
selected in the previous experiment for each dataset. Specifically, they were
set to $J=3$, learning rate to 0.5 and subsampling to 1 for {\it Tic-tac-toe},
to $J=1$, learning rate to 0.5 and subsampling to 0.75 for {\it Spambase}
and to $J=4$, learning rate to 0.5 and subsampling to 1 for {\it Energy-Heating}
for all partitions. 
Note that for each sequence of GBNN, only one model is trained. For NN, 200
independent models with 1 to 200 neurons need to be trained in order to obtain
the sequence.

From Fig~\ref{fig:acc_comparison}, it can be observed that the average test
accuracy of GBNN improves as more units are considered. 
More
importantly, we can observe that GBNN tends to stabilize with the number
of units. Hence, if the latter units are removed, the performance in the model
accuracy is not damaged to a great extent. For instance, in {\it Spambase},
if the number of units is reduced from $T=200$ to $T=100$ the model accuracy
only drops from $94.78\%$ to $\approx 94.63\%$. In {\it Tic-tac-toe} the
same size reduction does not reduce the accuracy of the model. This observation
also applies for the {\it Energy-Heating} regression task. This property
can be useful, as one can adopt a single model to different computational
requirements on the fly. The performance of NN could be higher than that of GBNN
at some stages (as shown for {\it Spambase}), however, the number of hidden
neurons to use have to be decided during train. In addition, the performace of
two neural networks with $T$ and $T+1$ trained on the same data present a higher
variance than a single GBNN model using $T$ and $T+1$. Hence, the performance of
the sequence of NN is not as monotonic as the sequence of GBNN.

\begin{figure}[t]
\centering
\begin{tabular}{@{}c@{}c@{}}
\subfloat[Power]{{\includegraphics[width=6.2cm]{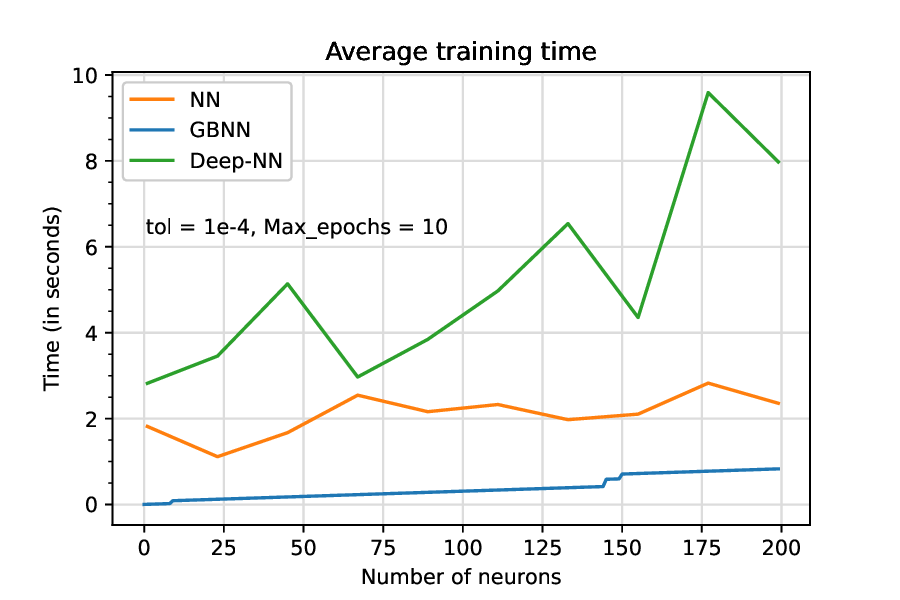}}}
\subfloat[Power]{{\includegraphics[width=6.2cm]{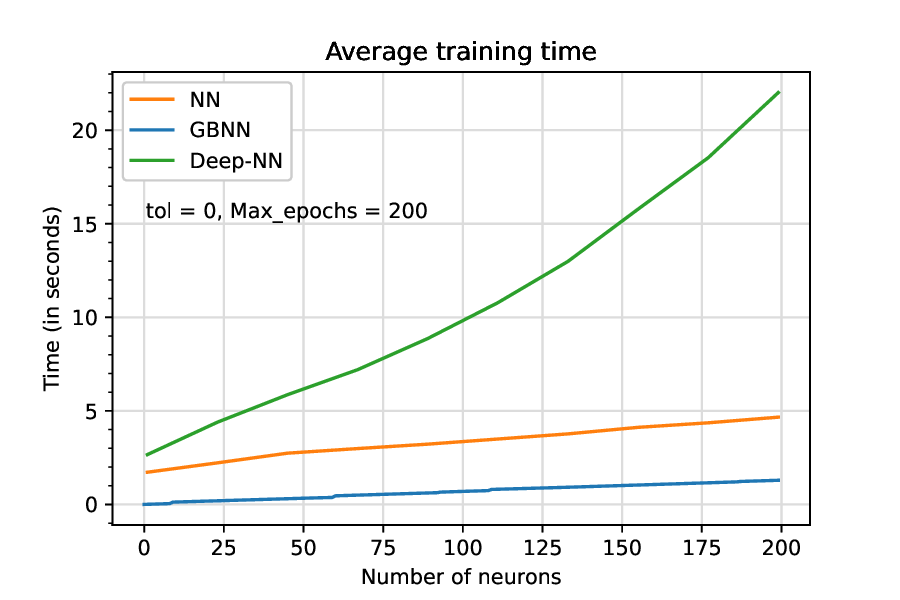}}}
\\
\subfloat[Indian Liver Patient]{{\includegraphics[width=6.2cm]{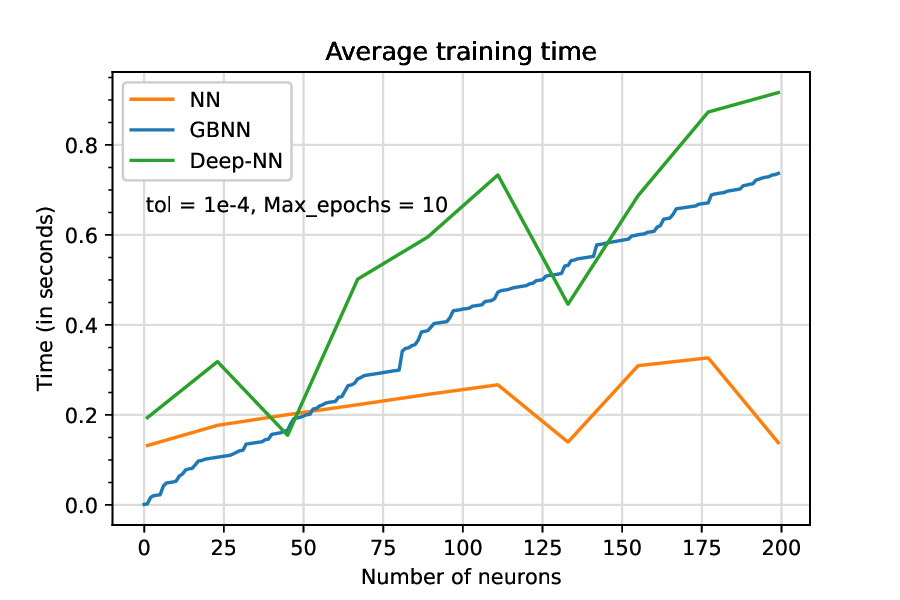}}}
\subfloat[Indian Liver Patient]{{\includegraphics[width=6.2cm]{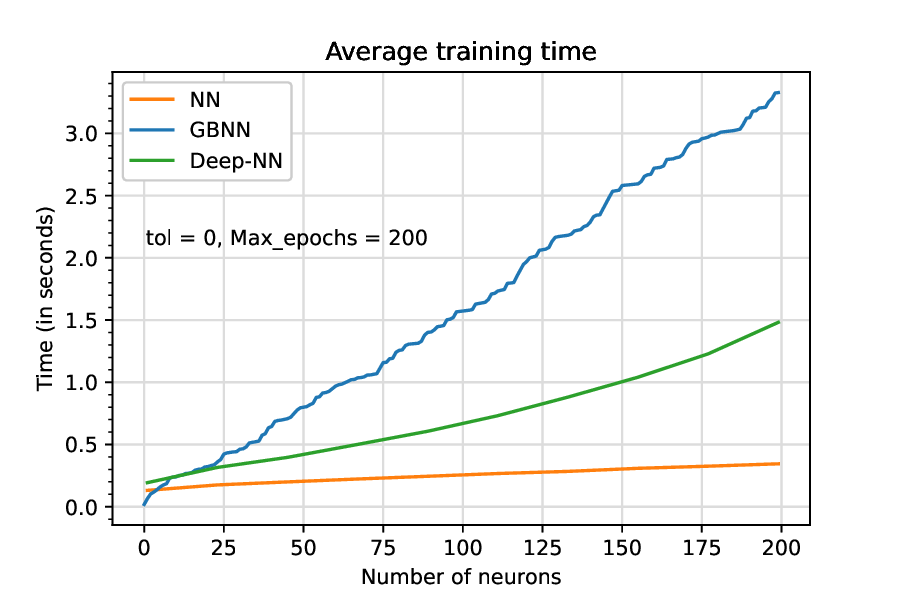}}}
\end{tabular}
\caption{Average training time (in seconds) for gradient boosted neural network
(Blue curve), standard neural network (orange), and deep neural network (green), 
for waveform (top row) and Indian liver patient (bottom row). 
In the left column, the default early stopping procedure was set to train the
networks of all models (tolerance is set to $1e-4$ and iterations with no change
to ten). In the right column, all networks are force to train during 200 epochs
(tolerance is set to 0 and max iterations without change to 200)
} \label{fig:Training_time}
\end{figure}

In order to compare the computational performance of the tested methods, 
the training time for GBNN, Deep-NN, NN--Adam, NN-L-BFGS and NN-SGD are
shown in Table~\ref{table:Training_time}.
For this experiment, we applied 10-fold cross-validation and computed the
average fit time of the final model that used the set hyper-parameters most
frequently selected in the cross-validation. 
The size for GBNN and the different networks was set to 200.
Deep-NN was built, as before with three layers of 100 neurons each. 
GBNN uses the solver L-BFGS. 
This experiment is done on CPU using AMD Ryzen 7 5800H 3.20 GHz processor.
As Table~\ref{table:Training_time} illustrates, the most computationally
efficient method is NN--SGD followed by NN--L-BFGS. In general, the
differences are rather contained among most methods and datasets. Some
exceptions include the Deep-NN with respect to the other methods, which could be
up to x20 slower (e.g. {\it Power} or {\it Energy-Heating}). The variance in the average
times among the different methods for each dataset is also due to the fact that
the default hyper-parameters for all methods include an early-stopping strategy in
which the training stops if the loss does not drop at least $0.0001$ in ten
training iterations. In order to visualize this effect,
in Figure~\ref{fig:Training_time} we show the average training time with respect
of the number of neurons in the hidden layers for GBNN, Deep-NN and NN-Adam
for two datasets (one favorable to GBNN and one favorable to NN). 
The left plot shows the results considering early stopping and
the right  plot forces the networks to train for 200 epochs. Note that the
results of GBNN are monotonic since a single model is used to obtain the
training time sequence. For Deep-NN and NN-Adam, a different model is trained
every 20 hidden neurons, which explains the peaks in the curves especially when
early stopping is active (left plots). When the models are forced to train for
200 epochs, the evolution of the shallow models show a clear linear
complexity with respect to the number of hidden units. The deep model shows a
quadratic complexity in accordance with the growth of the number of weights. 
In addition, the training time performance of GBNN
is also affected by the model setting. This can account for the favorable and
unfavorable results for GBNN shown in Figure~\ref{fig:Training_time}. In
particular, in some exploratory experiments we carried out using different learning
rates illustrated that the higher the learning rate, the slower the training
time specially for classification.

\begin{table*}[t!]
\centering
\resizebox{0.95\textwidth}{!}{%
\begin{tabular}{lccccc}
\hline
\textbf{Dataset} & \textbf{GBNN} & \textbf{Deep-NN}	& \textbf{NN--Adam} 	& \textbf{NN--L-BFGS} & \textbf{NN--SGD} \\
\hline
{\bf Binary classification}\\
Australian Credit Approval & 1.105 & 0.236 & \cellcolor{yellow!47}  0.093   & 0.614 & 0.359 \\
Banknote & 0.454  &  0.391 & 0.515 & \cellcolor{yellow!47} 0.065 & 1.055 \\
Breast cancer &  0.165  & 0.295 &  0.305 & 0.510 &\cellcolor{yellow!47}  0.054\\
Diabetes & 0.882  &  0.253 & 0.698 & 0.619 & \cellcolor{yellow!47} 0.221 \\
German Credit Data &  \cellcolor{yellow!47} 0.129 & 0.187 & 0.208 & 0.130 & 0.524 \\
Hepatitis & 0.316  & 0.060 & \cellcolor{yellow!47}  0.023 & 0.247 & 0.046\\
Indian Liver Patient & 0.557 &  0.220  & 0.288 & \cellcolor{yellow!47} 0.073 & 0.400 \\
Ionosphere & 0.791 & 0.618 &  0.431 & \cellcolor{yellow!47} 0.085 & 0.400 \\
MAGIC Gamma Telescope &  5.273 & 37.267 & 6.177 & 12.931 &\cellcolor{yellow!47}  5.038 \\
Spambase & 2.240  & 2.461  &  0.903 & 5.057 & \cellcolor{yellow!47} 0.639\\
Sonar & 0.568  & 0.581 &   0.327 & \cellcolor{yellow!47} 0.166 & 0.307\\
Tic-tac-toe &  0.732 & 1.215 & 1.011 & \cellcolor{yellow!47} 0.413 & 0.956 \\
\hline
{\bf Multi-class classification}\\
Digits &  0.976 & 1.095 & 1.016 & \cellcolor{yellow!47} 0.463 & 2.369 \\
Iris & 1.288  & 0.321  &  0.108 & 0.142 & \cellcolor{yellow!47} 0.102 \\
Vehicle &  \cellcolor{yellow!47} 0.256  & 0.519 &  0.269 & 0.549 & 0.391 \\	
Vowel &  1.088  & 1.992  & 1.419 & \cellcolor{yellow!47} 0.578 & 1.085 \\
Waveform &  3.896 & 8.676  & 5.452 & \cellcolor{yellow!47} 3.644 & 5.146 \\
Wine & 0.116 & 0.348 & \cellcolor{yellow!47}  0.022 & 0.071 & 0.041 \\
\hline
{\bf Regression}\\
Boston Housing & 0.636 & 1.026 &  0.468 & 0.344 & \cellcolor{yellow!47} 0.030 \\ 
Concrete &  0.453 &  0.650 & 0.672 & 0.673 & \cellcolor{yellow!47} 0.045 \\
Energy-Cooling &  0.077 & 0.768 & 0.632 & 0.573 & \cellcolor{yellow!47} 0.037 \\
Energy-Heating &  0.083 &  1.023 & 0.643 & 0.488 & \cellcolor{yellow!47} 0.037 \\
Power &  \cellcolor{yellow!47} 0.198  & 4.093 & 1.709 & 1.330 & 0.321 \\
Wine quality-red & 1.192 & 1.748 &  0.865 & 0.995 & \cellcolor{yellow!47} 0.450 \\ 
Wine quality-white &  1.148 & 1.953 & 2.647 & 3.150 & \cellcolor{yellow!47} 0.220 \\ 
\hline
\end{tabular}\hspace*{-25pt}
}
\caption{Total training time in seconds for gradient boosted neural network
(GBNN), deep neural network, and neural networks with Adam solver} 
\label{table:Training_time}
\end{table*}

\subsection{Experiment with large datasets and deep models}
For the second batch of experiment, three-layer deep dense network and CNNs were
used. In order to compare the deep models with the proposed method, we followed
a transfer learning approach. For this, once the deep models are trained, the
last dense layer and the output layer of the CNN and Deep-NN are removed. Then,
the weights of the first layers are frozen. Finally, a GBNN model is trained
linked to the frozen layers using the same training instances. We term this
model as Deep-GBNN. For {\it Covertype} and {\it Poker Hand}, a three-layer deep dense
neural network with 100 hidden networks per layer were trained for 200 epochs.
The activation functions are ReLu for the internal layers and
Sigmoid for the output layer. The solver SGD was applied. For {\it CIFAR-10} and
{\it MNIST}, a CNN was trained for 200 epochs on the training set. The CNN includes
the following layers: two convolutions(32), max-pooling (2x2), dropout(0.2), two
convolutions(64), max-pooling (2x2), dropout(0.3), two convolutions(128),
max-pooling (2x2), dropout(0.4), two dense(128), and one output dense layer for
ten classes. The activation functions are ReLu for the internal layers and
SoftMax for the output layer. For these datasets, a single partition train/test
was carried out. The dataset is divided into a training and a test set as
defined by the dataset: 50000 train and 10000 test for {\it CIFAR-10}; 60000
train and 10000 test for {\it MNIST} and 25010/1000000 for {\it Poker}. For {\it
Covertype}, a random stratified partition of 70\%-30\% was done. The optimum
hyper-parameters configuration was measured using within-train 5-fold
cross-validation, except for {\it Covertype} where a 2-fold cross-validation was
used. For Deep-GBNN, the values for the grid search were in the following
ranges: $[0.1, 1]$ for the learning rate, $[0.25, 1]$ for subsample and $[1,15]$
for the step size. Also, the solver for the GBNN method was set to Adam, due to
better performance in dealing with high-dimensional datasets. In addition, we
run the final experiment using pre-trained networks. For that, we
used the InceptionV3 \cite{szegedy2016rethinking} and VGG16
\cite{simonyan2014very} models pre-trained on the {\it imagenet} dataset
\cite{deng2009imagenet}. These models were loaded and subsequently fine-tuned
on the {\it CIFAR-10} dataset using the default train partition composed of
$50,000$ instances. They are validated on the remaining $10,000$ test instances.
Finally, we froze the weights of VGG16 and InceptionV3 and replaced their last 
dense layer with a optimized GBNN and NN with 500 units, in which the
optimization had done using grid search using train. Finally, the accuracy of
NN and GBNN is estimated on the test set.

The Table~\ref{table:results_CNN} presents the achieved accuracy for four
datasets in the test set (first two columns) and the estimation of the
generalization accuracy obtained in the in-train cross-validation (last two
columns).
The best in-train and test results are highlighted with a yellow background.
The results show uneven performance for the different dataset. 
The proposed method manages to improve the performance of
the deep models in {\it MNIST}, {\it Cover type} and {\it Poker hand} by 0.06,
1.76 and 0.12 percentage points respectively. 
In {\it CIFAR-10}, performance drops by 0.03\%. Even if the differences are in
general marginal, an interesting aspect is that the in-train performance can be
used to select the best model on each dataset except in {\it MNIST}. Although
the differences between method in this dataset both in-train and in test are
negligible.
Finally, the performances of the GBNN and NN models trained on the output of
InceptionV3 and VGG16 for {\it CIFAR-10}, and the fine-tuned transfer learning
models as well are shown in Table~\ref{table:results_pre_trained}. The
application of GBNN on top of the pre-trained and fine-tuned models, gained 
0.08 and 0.07 percent points in accuracy with respect to InceptionV3 and VGG16 
respectively. The results of NN are 0.08 percentage points worse and 0.08 better
than the deep fined-tuned models. The accuracy improvements of using GBNN on top
of the deep models is small but as it is the additional computational training 
cost. The classification computational cost remains the same.

\begin{table*}[h]
\centering
\resizebox{.9\textwidth}{!}{%
\begin{tabular}{lcccc}
\hline
 & \multicolumn{2}{c}{\textbf{Test estimation}}  & \multicolumn{2}{c}{\textbf{In-train estimation}}\\
\hline
\textbf{Dataset} & \textbf{Deep-GBNN} & \textbf{Deep-CNN}  & \textbf{Deep-GBNN}  & \textbf{Deep-CNN} \\
CIFAR-10 & 84.09 &  \cellcolor{yellow!47}  84.12  & 82.77 & \cellcolor{yellow!47} 83.04  \\
MNIST & \cellcolor{yellow!47} 99.52  & 99.46 & 99.35 & \cellcolor{yellow!47} 99.42\\
\hline
\textbf{ } & \textbf{Deep-GBNN} & \textbf{Deep-NN} & \textbf{Deep-GBNN}  & \textbf{Deep-NN} \\

CoverType & \cellcolor{yellow!47} 90.00 & 88.24 & \cellcolor{yellow!47} 87.53 & 83.45\\
Poker Hand & \cellcolor{yellow!47} 99.45 &  99.33 &  \cellcolor{yellow!47} 98.57 & 97.72\\
\hline
\end{tabular}\hspace*{-25pt}
}
\caption{Average generalization performance (first two columns) and in-train
performance (second two columns) for deep gradient boosted neural networks
(Deep-GBNN) and Deep Convolutional Neural Networks (Deep-CNN).}

\label{table:results_CNN}
\end{table*}

\begin{table*}[h]
\centering
\resizebox{0.9\textwidth}{!}{%
\begin{tabular}{lccccc}
\hline
& {\textbf{Fine-tuned}} &  \multicolumn{2}{c}{ \textbf{GBNN}}  & \multicolumn{2}{c}{ \textbf{NN}}\\
\hline
\textbf{Pre-trained model} & \textbf{Test} & \textbf{Test} & \textbf{In-train}  & \textbf{Test}  &  \textbf{In-train}\\

InceptionV3 & 93.12 & \cellcolor{yellow!47}93.20 & \cellcolor{yellow!47}99.99 & 93.04 & 99.98 \\
VGG16 &  92.92 &  92.99 & \cellcolor{yellow!47}99.98 & \cellcolor{yellow!47} 93.00 & 99.98 \\
\hline
\end{tabular}\hspace*{-25pt}
}
\caption{The generalization performance of the fine-tuned transfer learning models for each model (InceptionV3 and VGG16). And generalization and in-train performance of a one-layer GBNN and NN-trained models on the output of pre-trained models on the CIFAR-10.}

\label{table:results_pre_trained}
\end{table*}

\section{Conclusions}
In this paper, we present a novel iterative method to train a neural network
based on gradient boosting. The proposed algorithm builds at each step a
regression neural network with one or few ($J$) hidden unit(s) fitted to the
data residuals. The weights of the network are then updated with a
Newton-Raphson step to minimize a given loss function. Then, the data residuals
are updated to train the model of the next iteration. The resulting $T$
regressors constitute a single neural network with $T \times J$ hidden units. In
addition, the formulation derived for this works opens the possibility to create
gradient boosting ensembles composed of base models different from decision
trees, as done in previous implementations of gradient boosting.

In the analyzed problems, the proposed method achieves a generalization accuracy
that converges with the number of combined regressors (or hidden units). This
quality of the proposed method allows us to use the combined model fully or
partially by deactivating the units in order inverse to their creation,
depending on classification speed requirements. This can be done on the fly
during the test. In addition, we showed that the training complexity is
equivalent to that of training a network with a standard solver.

The proposed method tested on a variety of classification, regression and image
processing tasks. The results show a performance favorable to the proposed
method in general. The proposed approach showed the best overall average rank
in the tested classification and regression problems with statistically
significant differences with respect to SGD and L-BFGS approaches. In addition,
for deep models a transfer learning approach was followed and the results were
favorable to the proposed method in some of the tasks although the differences
were small.
Notwithstanding, the proposed iterative training procedure opens novel
alternatives for training neural networks. This particularly evident for
regression tasks where the proposed method achieved the best result in most of
the analyzed datasets.

Machine learning is becoming a fundamental piece for the success of more and
more applications every day. Some examples of novel applications include
bioactive molecule prediction \cite{babajide_2016_molecule}, renewable energy
prediction \cite{torres_2017_windsolar}, classification of galactic sources
\cite{mirabal_2016_galactic}, or agriculture area for mapping soil contamination
\cite{jia2021mapping}. It is of capital importance to find algorithms that can
efficiently handle complex data. 
Ensemble methods are very effective at improving the generalization accuracy of
multiple simple models  \cite{classifiersarticle,caruana06empirical} or even
complex models such as MLPs \cite{schwenk_2000_boostingnn} or DeepCNNs
\cite{Moghimi2016}. 

In recent years, gradient boosting \cite{gradientboosting,friedman2000additive}, a fairly old
technique has gained much attention, specially due to the novel and
computationally efficient
version of gradient boosting called eXtreme Gradient Boosting or XGBoost
\cite{XGBoost}. Gradient boosting builds a model as an additive expansion of
regressors to gradually minimize a given loss function. When gradient boosting
is combined with several stochastic techniques, as bootstrapping or feature
sampling from random forest \cite{randomforests}, its performance generally
improves \cite{stochastic_gb}. In fact, this combination of randomization
techniques and optimization has placed XGBoost among the top contenders in
Kaggle competitions \cite{XGBoost} and provides excellent performance in a
variety of applications as in the ones mentioned above. Based on the success of
XGBoost, other techniques have been proposed like CatBoost
\cite{prokhorenkova2018catboost} and LightGBM \cite{ke2017lightgbm}, which
propose improvements in training speed and generalization performance. 
More details
about these methods can be seen in the comparative analysis of Bentéjac
et al. \cite{bentejac_2020_comparative}.
Other type of widespread boosting algorithm is AdaBoost
\cite{freund1997decision}, initially developed for binary classification and
then for multi-class classification (AdaBoost-SAMME)
\cite{hastie2009multi} and regression \cite{drucker1997improving}. 

On the other hand, convolutional deep architectures have shown outstanding
performances especially with structured data such as images, speech, etc.
\cite{lecun2015deep,schmidhuber2015deep}. However, in the context of tabular data,
ensembles of classifiers or simple MLPs are generally more effective than
convolutional deep neural networks \cite{comparison2017zhang}. The objective of this study is to combine the stage-wise optimization of
gradient boosting into the training procedure of the last layers of a
neural network. The
result of the proposed algorithm is an alternative for training a single neural
network (not an ensemble of networks).

Several related studies propose hybrid algorithms that, for instance, transform
a decision forest into a single neural network \cite{welbl2014casting,Biau2018}
or that use a deep architecture to train a tree forest
\cite{Kontschieder_2015_ICCV}. In \cite{welbl2014casting},
it is shown that a pre-trained tree forest can be cast into a two-layer neural
network with the same predictive outputs. First, each tree is converted into a
neural network. To do so, each split in the tree is transformed into an
individual neuron that is connected to a single input attribute (split
attribute) and whose activation threshold is set to the split threshold. In this
way, and by a proper combination of the outputs of these neurons (splits) the
network mimics the behavior of the decision tree. Finally, all neurons are
combined through a second layer, which recovers the forest decision. The weights
of this network can be later retrained to obtain further improvements
\cite{Biau2018}. In \cite{Kontschieder_2015_ICCV}, a
decision forest is trained jointly by means of a deep neural network that learns
all splits of all trees of the forest. To guide the network to learn the splits
of the trees, a procedure that trains the trees using back-propagation, is
proposed. The final output of the algorithm is a decision forest whose
performance is remarkable in image classification tasks.

In other related line of work \cite{nitanda2018functional,huang2018learning},
boosting is applied to the construction of Deep Residual Learning models
\cite{he2016deep}. In \cite{nitanda2018functional}, a novel ResNet weight
estimation model is proposed by generalizing the boosting functional gradient
minimization \cite{mason1999boosting} to the feature extraction space of the
network. The work presented in \cite{huang2018learning} also builds
layer-by-layer a ResNet boosting over features, however, it is based on a
different boosting framework \cite{freund1997decision}.  The model (called
BoostResNet) works by learning a linear classifier on the output of each
residual network block to build an ensemble of shallow blocks. One important
advantage of BoostResNet over standard ResNet is its lower computational
complexity although the reported performance is not consistently better with
respect to ResNet. In contrast, the current proposal method, based on
\cite{gradientboosting}, builds a simple shallow network in-width rather than
complex model in-depth and shows very good performance in tabular datasets with
respect to standard back-propagation training methods. Furthermore, our proposal
can adapt on the fly to the use of a reduced number of hidden neurons.

Another model that resembles the idea proposed in this paper --yet with a 
different optimization process, final model and objective-- was presented in
\cite{bengio2005convex}. 
They propose a convex optimization algorithm for training a neural network that
theoreticallly could reach the global optimum although its exact implementation
is only feasible for a very low number of input features. In
order to reach the global optimum they control the number of hidden neurons of
the model by adding one neuron at a time to the network and by
including a $L^1$ regularization on the top layer. The proposed idea is a
stepwise algorithm as all weights of the network are optimized at each
iteration. This is done in three optimization steps. First, a new neuron (i.e.
linear model) is added and trained on a weighted loss function similarly to
Adaboost. This weighted loss can only be solved exactly for a very low number of
input features. Then, the output layer, and potentially all input weights, are
optimized using the proposed convex formulation. Finally, the output weights are
regularized to reduce the complexity of the network. This final step sets to
zero some of the output weights to effectively remove the corresponding neurons.
The algorithm is tested on one simple 2D problem in order to assess the validity
of the global optimum approach.

In this paper, we propose a combination of ensembles and neural networks that is
somehow complementary to the work of
\cite{Kontschieder_2015_ICCV}, which is a single neural
network that is trained using an ensemble training algorithm.
Specifically, we propose to train a neural network iteratively as an additive
expansion of simpler models. The algorithm is equivalent to gradient boosting:
first, a constant approximation is computed (assigned to the bias term of the
neural network), then at each step, a regression neural network with a single
(or very few) neuron(s) in the hidden layer is trained to fit the residuals of
the previous models. All these models are then combined to form a single neural
network with one hidden layer. This training procedure provides an alternative
to standard training solvers (such as Adam, L-BFGS or SGD) for training a neural
network.
Other works related to the optimization and convergence of Adam have been
recently proposed
\cite{reddi2019convergence,defazio2021adaptivity,chen2018closing}. They showed
that its convergence can be improved in the context of high dimensional complex
image classification tasks. However, their focus is mainly in deep models for
image classification tasks. 

In addition, the proposed method has an additive neural architecture in
which the latest computed neurons contribute less to the final decision. This
can be useful in computationally intensive applications as the number of active
models (or neurons) can be gauged on the fly to the available computational
resources without a significant loss in generalization accuracy. The proposed
model is tested on multiple classification and regression problems, as well as
in conjunction with deep models posed as transfer learning problems. 
These experiments show that the proposed method for training the last layers of
a neural network is a good alternative to other standard methods.

The paper is organized as follows: Section 2 describes gradient boosting and how
to apply it to train a single neural network; In section 3 the results of
several experimental analysis are shown; Finally, the conclusions are summarized
in the last section.

In this paper, we present a novel iterative method to train a neural network
based on gradient boosting. The proposed algorithm builds at each
step a regression neural network with one or few ($J$) hidden unit(s) fitted to
the data residuals. The weights of the network are then updated with a
Newton-Raphson step to minimize a given loss function. Then, the data residuals
are updated to train the model of the next iteration. The resulting $T$
regressors constitute a single neural network with $T \times J$
hidden units. In addition, the formulation derived for this works opens the
possibility to create gradient boosting ensembles composed of base models
different from decision trees, as the all previous implementations.

In the analyzed problems, the proposed method achieves
a generalization accuracy that converges with the number of combined regressors
(or hidden units). This quality of the proposed method allows us to use
the combined model fully or partially by deactivating the units in order
inverse to their creation, depending on classification speed requirements. This
can be done on the fly during test.

The proposed method has been tested on a variety of classification and
regression tasks. The results show a performance favorable to the proposed
method in general although the overall differences are not statistically
significant with respect to solvers Adam and L-BFGS (they are with respect to
SGD). Notwithstanding, the proposed iterative training procedure opens novel
alternatives for training neural networks.
This is particularly evident for regression tasks where the proposed method
achieved the best result in most of the analyzed datasets.

\bibliographystyle{plain}
\bibliography{refs}
\end{document}